\documentclass[conference]{IEEEtran}
\IEEEoverridecommandlockouts

\usepackage{cite}
\usepackage{amsmath,amssymb,amsfonts}
\usepackage{algorithmic}
\usepackage{graphicx}
\usepackage{textcomp}
\usepackage{xcolor}
\usepackage{booktabs}
\usepackage{tabularx}
\usepackage{multirow}
\usepackage{lipsum}  
\usepackage{xcolor}

\def\BibTeX{{\rm B\kern-.05em{\sc i\kern-.025em b}\kern-.08em
    T\kern-.1667em\lower.7ex\hbox{E}\kern-.125emX}}
\begin{document}

\title{Enhancing Vehicle Aerodynamics with Deep Reinforcement Learning in Voxelised Models \\
\thanks{979-8-3503-7565-7/24/\$31.00 ©2024 IEEE}
}

\author{\IEEEauthorblockN{Jignesh Patel}
\IEEEauthorblockA{\textit{Department of Networks and} \\ \textit{Digital Media,} \\
\textit{Kingston University,}\\
London, UK \\
jignesh.patel@kingston.ac.uk}
\and
\IEEEauthorblockN{Yannis Spyridis}
\IEEEauthorblockA{\textit{Department of Networks and} \\ \textit{Digital Media,} \\
\textit{Kingston University,}\\
London, UK \\
y.spyridis@kingston.ac.uk}
\and
\IEEEauthorblockN{Vasileios Argyriou}
\IEEEauthorblockA{\textit{Department of Networks and} \\ \textit{Digital Media,} \\
\textit{Kingston University,}\\
London, UK \\
vasileios.argyriou@kingston.ac.uk}
}

\maketitle

\begin{abstract}
Aerodynamic design optimisation plays a crucial role in improving the performance and efficiency of automotive vehicles. This paper presents a novel approach for aerodynamic optimisation in car design using deep reinforcement learning (DRL). Traditional optimisation methods often face challenges in handling the complexity of the design space and capturing non-linear relationships between design parameters and aerodynamic performance metrics. This study addresses these challenges by employing DRL to learn optimal aerodynamic design strategies in a voxelised model representation. The proposed approach utilises voxelised models to discretise the vehicle geometry into a grid of voxels, allowing for a detailed representation of the aerodynamic flow field. The Proximal Policy Optimisation (PPO) algorithm is then employed to train a DRL agent to optimise the design parameters of the vehicle with respect to drag force, kinetic energy, and voxel collision count. Experimental results demonstrate the effectiveness and efficiency of the proposed approach in achieving significant results in aerodynamic performance. The findings highlight the potential of DRL techniques for addressing complex aerodynamic design optimisation problems in automotive engineering, with implications for improving vehicle performance, fuel efficiency, and environmental sustainability.
\end{abstract}


\section{Introduction}

In the constantly evolving field of vehicle performance, aerodynamic design optimisation plays a key role. While mainly driven by the high-performance car and motorsport industries, optimising aerodynamic design is also relevant in everyday vehicles. Good aerodynamics can affect several aspects, such as fuel efficiency and interior noise reduction, thus improving the driving experience. The rapid increase in electric vehicle (EV) adoption and the impending prohibition on petrol vehicles within the European Union, further motivates aerodynamic optimisation, as EVs have a greater emphasis on energy efficiency due to range considerations.

Computational fluid dynamics (CFD) has been the predominant driving force in aerodynamic design for decades. Recently, surrogate models have been investigated to approximate the behaviour of computationally intensive processes associated with CFD simulations, resulting in more efficient optimisation methods. These involve parametric approaches, such as polynomial regression and kriging models \cite{wang2010aerodynamic}, or non-parametric methods, such as radial basis functions \cite{urquhart2020surrogate}, and artificial neural networks (ANNs) \cite{sun2019review}.

The application of ANNs in aerodynamic design optimisation encompasses additional aspects, including aerodynamic data modeling and parameter estimation \cite{sun2019review}, thus leading to the capture of complex relationships between the design parameters and aerodynamic performance. This allows the representation of non-linear relationships in high-dimensional spaces, without explicitly requiring mathematical models. ANNs in the form of convolutional neural networks, and generative adversarial networks (GANs) have been applied in the context of aerodynamic optimisation, in applications ranging from airfoil design in aircraft to wind turbine balance \cite{chen2020multiple,cao2019novel,du2020b}.

Based on the characteristics of ANNs and their ability to model aerodynamic parameters, this work proposes the use of deep reinforcement learning (DRL) to enhance the flexibility within the aerodynamic design process. The integration of DRL introduces adaptability to changing objectives or constraints within the design problems. DRL algorithms enable the exploration of solution spaces in a more comprehensive manner, which is particularly beneficial in multi-objective optimisation scenarios where conflicting design goals must be balanced \cite{li2020deep}. Moreover, DRL methods offer improved data efficiency by utilising past experiences to guide future decisions, thereby reducing the computational cost of exploring large design spaces \cite{yang2021exploration}. Additionally, DRL models have the ability to generalise to new designs by learning underlying patterns \cite{vithayathil2020survey}, enhancing their applicability across diverse aerodynamic applications. Therefore, the integration of DRL into aerodynamic design optimisation utilises the strengths of the underlying neural networks, while introducing adaptability, efficiency, and generalisation capabilities essential for tackling complex design challenges in automotive engineering.

The remainder of this paper is organised as follows: Section \ref{sec:background} explores the background on voxelisation and aerodynamic optimisation. Section \ref{sec:methodology} presents the methodology of our approach. Section \ref{sec:evaluation} presents the outcomes of the DRL  process and highlight the implications of the study. Finally, section \ref{sec:conclusion} summarises key findings and projects future work.

\section{Background}
\label{sec:background}

\subsection{3D Model Voxelisation}
Voxelisation is the process of converting a 3D model into a set of volumetric elements called voxels, which represent the spatial occupancy of the model. The technique has gained significant attention due to its wide range of applications. In CFD specifically, voxelisation is used to represent complex geometries for fluid flow simulations. The use of voxel-based 3D models allows for the systematic decomposition of geographical space into cuboid volumetric elements, enabling the modeling of continuous phenomena such as fluid flow \cite{aleksandrov2021voxelisation}. 

Beyond CFD, voxelisation has found applications in 3D city modeling \cite{nourian2016voxelization}, virtual/augmented reality \cite{yang2022vox}, and 3D printing \cite{telea2011voxel}. Voxel-based representations have been instrumental in urban planning and design, as well as in the development of immersive virtual environments \cite{lindlbauer2018remixed}. In addition, image-to-voxel translation networks have utilised the use of implicit functions and GANs for voxel-based 3D model representation, enabling the reconstruction and generation of high-quality 3D objects \cite{kniaz2019generative}.

Recent advances in voxelisation techniques have focused on indoor reconstruction \cite{hubner2021automatic}, addressing the challenges of manually reconstructing digital models of building interiors from unstructured triangle meshes. Such methods are capable of deriving semantically-enriched and geometrically completed indoor models in voxel representation without requiring planar or aligned room surfaces, clear vertical subdivisions, or distinct floor levels. Additionally, recent studies have explored voxel-based three-view hybrid parallel networks for 3D shape classification \cite{cai2021voxel}. This approach preserves spatial information by obtaining depth projection views from multiple angles, ensuring better recognition accuracy. Voxel-based feature engineering has also been investigated towards point cloud data processing, aimed at improving the characterisation of point clusters and supporting both supervised or unsupervised classification tasks \cite{poux2019voxel}.

\subsection{Aerodynamic Optimisation}
Aerodynamic optimisation involves the systematic improvement of aerodynamic designs to achieve specific objectives and performance indices. Gradient-free algorithms, including metaheuristics such as genetic algorithms \cite{saleem2020aerodynamic} and particle swarm optimisation \cite{li2020improving}, have been utilised for CFD optimisation, however they are generally outperformed by gradient-based methods in high-dimensional problems \cite{lyu2014benchmarking}. The latter calculate the gradients of objective functions to direct the optimisation, leading to a complexity directly associated with the range of the design variables. While gradient-based methods can effectively find local optima, they are not often capable of obtaining the global optimal solutions \cite{skinner2018state}.

Surrogate-based optimisation has been proposed to reduce the computational cost in aerodynamic optimisation, by lowering the required numerical simulations through simpler mathematical relationships. A kriging surrogate model combined with PSO can greatly reduce aerodynamic interference, optimising the associated characteristics by solving Navier–Stokes equations \cite{li2013aerodynamic}. In order to address scalabillity issues at high-dimensional cases, a snapshot active subspace method can be adopted to reduce the input dimensions of the aerodynamic shape optimisation problem, thus reducing computational cost and maintaining feasibility \cite{li2019surrogate}. 

State-of-the-art methods in aerodynamic optimisation have seen advancements in the use of deep learning techniques to calculate the gradient flow in the optimisation framework, doubling the efficiency compared to traditional methods \cite{li2020efficient}. These methods cal also reduce the number of design variables and computational costs by learning from historical designs, while also improving prediction accuracy through handling larger volumes of training data. Furthermore, by utilising advanced models such as physics-informed neural networks to address off-design constraints and discontinuous aerodynamic functions, they ultimately contribute to more efficient and effective optimisation architectures \cite{li2022machine}.

\subsection{Reinforcement Learning in Automotive Engineering}
Reinforcement learning has proven to be a valuable tool in aerodynamic optimisation, playing a significant role in shaping various components. One of the key strengths of RL lies in its capacity to acquire complex skills through trial-and-error, making it particularly well-suited for tasks demanding intelligence and experience, such as aerodynamic design \cite{li2021learning}. By integrating RL with deep learning into optimisation frameworks, decision-making processes can be significantly enhanced, achieving superior outcomes compared to conventional methods \cite{arulkumaran2017brief}. 

RL with transfer learning has demonstrated its utility in extracting and leveraging experiences derived from semi-empirical methods, thereby leading to notable reductions in computational overhead in the optimisation of missile control surfaces \cite{yan2019aerodynamic}. Specifically, the Proximal Policy Optimisation (PPO) algorithm has been effectively employed in airfoil design tasks, directly optimising 2D Bezier curves through comprehensive evaluations of the fluid dynamics around the generated shapes \cite{viquerat2021direct}. Overall, the literature highlights the effectiveness of DRL techniques to achieve robust optimisation solutions, utilising parallel computational capabilities and transfer learning mechanisms to improve convergence rates. Nevertheless, challenges remain in navigating high-dimensional search spaces, expected to be addressed with future advancements in the field to further enhance optimisation methodologies \cite{garnier2021review}.

\section{Methodology}
\label{sec:methodology}

\begin{figure}[t]
	\centering
	\includegraphics[width=0.6\linewidth]{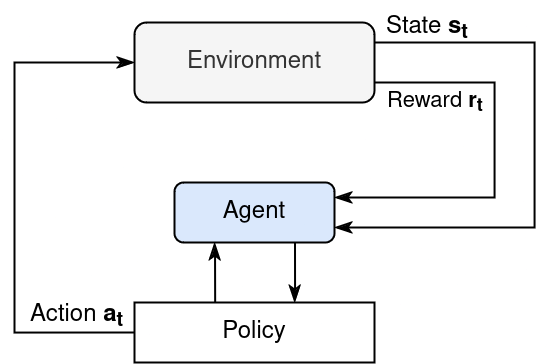}
	\caption{Interaction of an agent with the environment in reinforcement learning.}
	\label{fig:rl}
\end{figure}

Reinforcement learning is a machine learning technique that allows an agent to make decisions by interacting with the environment. RL operates on the principle of trial and error, where the agent receives feedback in the form of rewards or penalties based on the actions it takes, with the objective to maximise the cumulative reward over a specific time horizon. Through repeated experiences, the agent updates its decision-making strategy to favour actions that lead to greater rewards over time, thus learning an optimal policy. The key characteristics of RL are the environment, which provides states and rewards, and the agent, which selects actions and develops a policy, as illustrated in Figure \ref{fig:rl}. In RL the model learns by exploring and interacting with the environment, without relying on labelled or unlabelled data, as in supervised and unsupervised training respectively.

RL methods can follow the model-free or model-based approach. Model-free methods directly learn a policy without explicitly modelling the environment dynamics. While requiring significant amount of data to learn efficiently, they are often more flexible and adaptable to complex environments. Model-based methods on the other hand, learn a model of the environment through which they indirectly plan actions. In the context of optimising aerodynamic designs, model-free methods are preferred, since it is challenging to accurately model the environment dynamics. In addition, these methods can better adapt to complexities through direct interactions.

\subsection{Proximal Policy Optimisation}
DRL integrates deep neural networks with reinforcement learning algorithms to solve complex decision-making tasks. This study employs PPO, a DRL algorithm, which improves training stability and sample efficiency compared to traditional model-free policy gradient methods. PPO operates by iteratively updating the policy parameters to maximise the expected cumulative reward, while constraining the policy update to prevent large policy changes. The objective function is expressed as follows:

\begin{equation}
    L(\theta) = \min\left(r_t(\theta) \cdot \hat{A}_t, \text{clip}\left(r_t(\theta), 1 - \epsilon, 1 + \epsilon\right) \cdot \hat{A}_t\right),
\end{equation}

\noindent where $r_t(\theta)$ is the ratio of probabilities of actions under the updated ($\pi_{\theta}$) and current ($\pi_{\theta_k}$) policies:

\begin{equation}
    r_t(\theta) = \frac{\pi_{\theta}(a_t|s_t)}{\pi_{\theta_k}(a_t|s_t)},
\end{equation}

\noindent $L(\theta)$ is the objective function with respect to the policy parameters $\theta$, $\hat{A}_t$ is the advantage estimate at time step $t$, and $\epsilon$ is a hyperparameter that controls the degree of policy change.

\subsection{Voxelisation}
\begin{figure}[b]
	\centering
	\includegraphics[width=\linewidth]{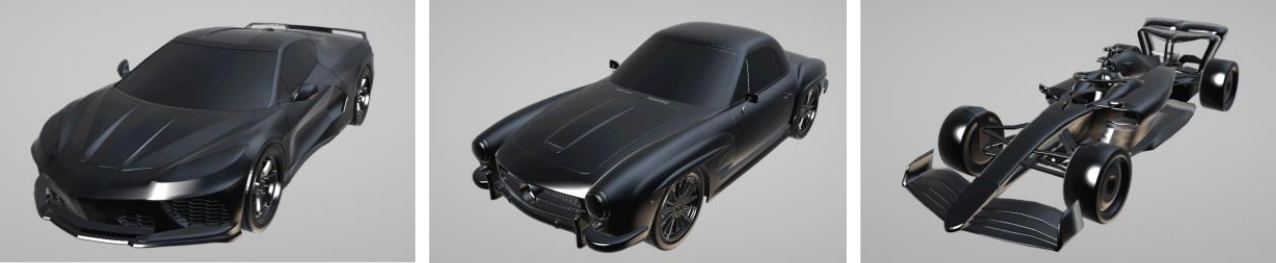}
	\caption{Original car prototypes.}
	\label{fig:models}
\end{figure}

\begin{figure}[b]
	\centering
	\includegraphics[width=\linewidth]{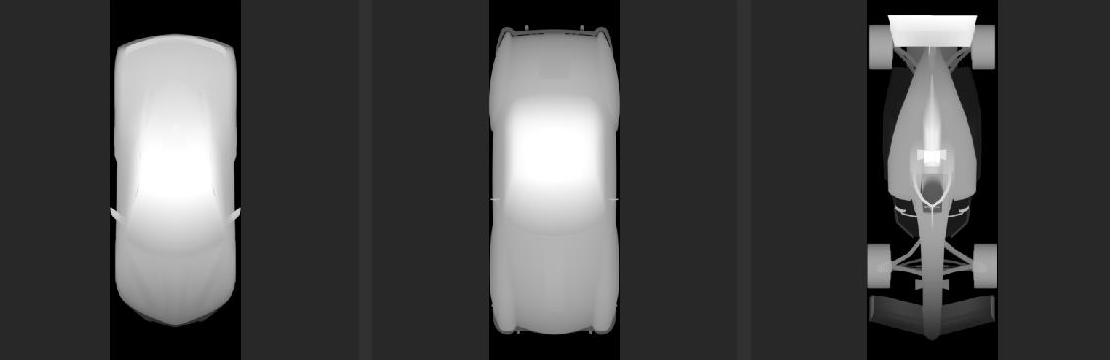}
	\caption{Generated height maps.}
	\label{fig:height_maps}
\end{figure}

\begin{figure}[b!]
	\centering
	\includegraphics[width=\linewidth]{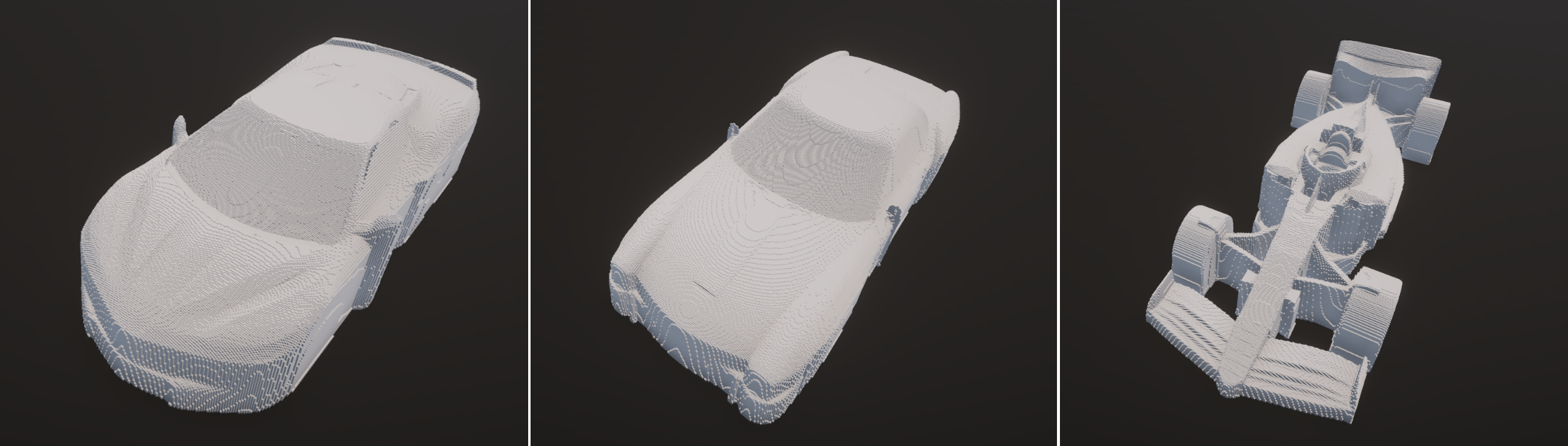}
	\caption{Voxelised models.}
	\label{fig:voxel_models}
\end{figure}

In the context of this work, voxelisation is a crucial step in the computational representation of 3D models, employed to convert the car designs to a format suitable for the simulation environment. Several models were acquired from CGTrader and processed in Blender to generate high resolution height maps, allowing to capture the elevation information of the prototypes. The height maps were then imported into Unity, where a voxel grid was constructed. Each voxel within this grid corresponds to an individual pixel in the height map, effectively discretising the representation into a 3D grid. The height value of each voxel is determined based on the corresponding pixel's elevation in the height map. By employing this process, the refined geometry of the car models is accurately translated into a voxelised representation, facilitating their integration into the simulation environment within Unity. It should be noted that the voxelised models are solid representations without internal voids or empty spaces. Figures \ref{fig:models}, \ref{fig:height_maps}, and \ref{fig:voxel_models} depict the original prototypes, the generated height maps, and the voxelised models respectively.

\subsection{Environment}
To emulate the wind tunnel environment, this study employs the physics simulation capability of Unity's Data-Oriented Technology Stack. The environment is simulated by introducing multiple spherical entities representing air particles, which are propelled towards the voxelised representations of the car models. These models involve three different vehicles, a Formula 1 (F1) car, a Duesen Bayern, and a Corvette C8. The air particles are subjected to physical forces, mimicking the behaviour of airflow within the wind tunnel environment. Throughout the simulation, the considered metrics include the drag force, kinetic energy, and voxel collision count. Consideration is also given to the summation of the voxel height values, as it affects how much the design changes. More specifically the metrics are defined as follows.

Drag is the force acting opposite to the relative motion of the car moving through the air. Reducing drag is crucial for improving fuel efficiency and increasing speed. The drag force is given by:

\begin{equation}
    F_{\text{drag}} = \frac{1}{2} \times \rho \times v^2 \times C_d \times A
\end{equation}

\noindent where $\rho$ is the density of the fluid, $v$ is the speed of the air particle object, $A$ is the cross-sectional area, and $C_d$ represents the drag coefficient. 

In the context of car aerodynamics simulations, kinetic energy is the energy possessed by the air particles due to their motion. Kinetic energy is relevant because it affects the vehicle's dynamics and interactions with the surrounding air. It is calculated as:

\begin{equation}
    KE = \frac{1}{2} \times m \times \upsilon^2
\end{equation}

\noindent where $m$ is the mass of the air particle and $\upsilon$ its velocity. Designs that maximise the kinetic energy of air particles, essentially accelerate the airflow around the vehicle and create a lower pressure zone, which can decrease the overall drag force.

\begin{figure}[b!]
	\centering
	\includegraphics[width=\linewidth]{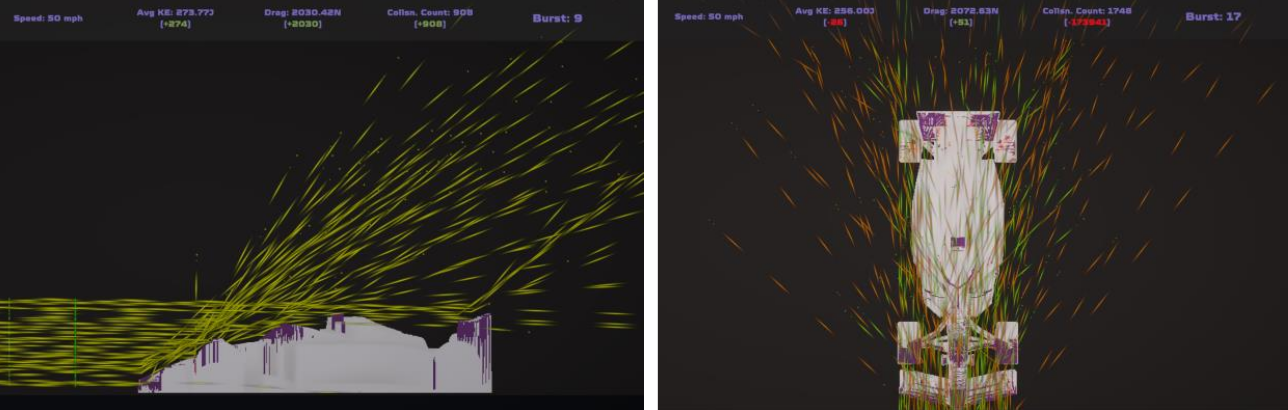}
	\caption{Aerodynamic simulation environment.}
	\label{fig:simulation}
\end{figure}

The voxel collision count is the number of times each air particle collides with a voxel. It is expressed as follows:

\begin{equation}
    C = \sum_{i=0}^{n} v_c \times \frac{1}{b \times b_c}
\end{equation}

\noindent where $v_c$ is the number of times an air particle collides with a voxel, $n$ the number of colliding
particles, $b$ is the burst count set, and $b_c$ represents the constant value for a base cycle count.

The heightmap sum variance involves the summation of the height values $h$ of all $n$ voxels in the heightmap, given by:

\begin{equation}
    H_s = \sum_{i=0}^{n} h
\end{equation}

During the agent-environment interaction, the PPO agent receives observations from the environment about the wind tunnel and the current design state of the car. These serve as inputs that guide the agent's decision-making process, enabling it to strategically select actions to modify the design, through the adjustment of the voxel height. Following the implementation of the modified design, the environment recalculates the aerodynamic metrics, which are then used to calculate the reward signal. This signal guides the agent's learning process and contributes to the refinement of its policy. Through this iterative refinement the metrics are minimised over time, thus facilitating the optimisation of the car design. Figure \ref{fig:simulation} illustrates a simulation instance.

\begin{table}[b!]
    \centering
    \caption{Training Parameters}
    \label{tab:params}
    {\renewcommand{\arraystretch}{1.4}
    \begin{tabularx}{0.9\linewidth}{|l|X|}
    \hline
    \textbf{Parameter} & \textbf{Value} \\
    \hline
    Batch Size & 1024 \\
    Replay Buffer Size & 10,240 \\
    Learning Rate & $3.0 \times 10^{-4}$ \\
    PPO Beta & $9.0 \times 10^{-3}$ (constant schedule) \\
    PPO Epsilon & 0.2 (linear schedule) \\
    PPO Lambda & 0.95 \\
    Training Epochs & 5 \\
    Neural Network Layers & 2 \\
    Hidden Units per Layer & 128 \\
    Memory Sequence Length & 64 \\
    Memory Size & 256 \\
    Max Training Steps & 5000 \\
    Time Horizon & 64 \\
    Summary Update Frequency & 1 \\
    Extrinsic Reward Signal Strength & 1.0 \\
    Discount Factor & 0.99 \\
    \hline
    \end{tabularx}
    }
\end{table}

\begin{table}[b!]
    \centering
    \caption{Simulation Configuration}
    \label{tab:sim_config}
    {\renewcommand{\arraystretch}{1.4}
    \begin{tabularx}{0.9\linewidth}{|l|X|}
    \hline
    \textbf{Parameter} & \textbf{Description} \\ 
    \hline
    Air Speed   & Speed of air in mph (10 to 120)         \\
    Air Density & Amount of air particles in simulation   \\
    Burst Count & Number of simulation runs per iteration \\
    Car Model   & Selected car model for simulation       \\
    \hline
    \end{tabularx}
    }
\end{table}

\subsection{Model training}

\begin{figure*}[t!]
	\centering
	\includegraphics[width=0.8\linewidth]{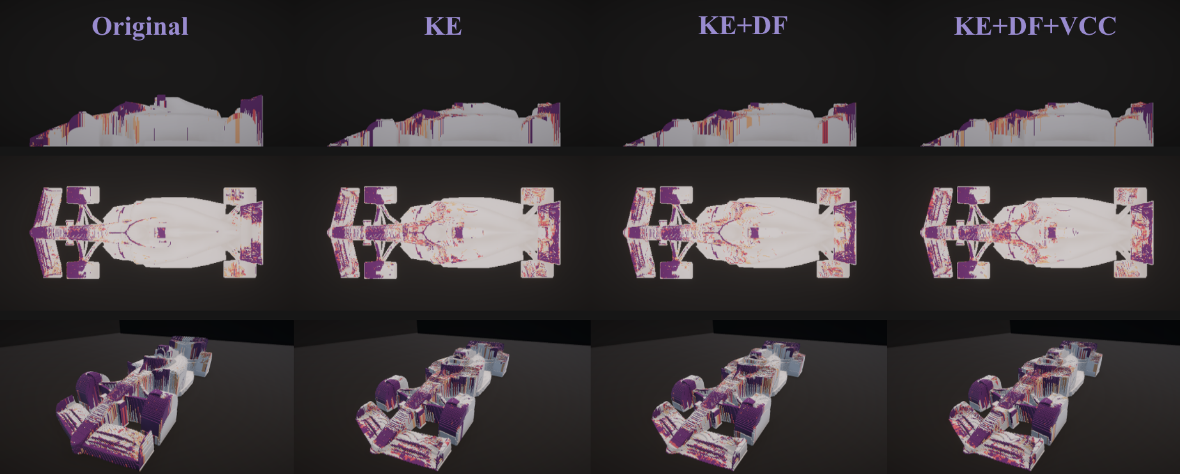}
	\caption{Collision heatmap of the F1 car before and after the optimisation, under different optimisation objectives.}
	\label{fig:collision_heatmap}
\end{figure*}

\begin{figure}[t!]
	\centering
	\includegraphics[width=0.7\linewidth]{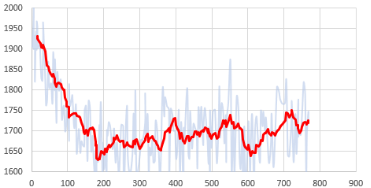}
	\caption{F1 model drag force throughout the training duration.}
	\label{fig:df}
\end{figure}

\begin{figure}[t!]
	\centering
	\includegraphics[width=0.7\linewidth]{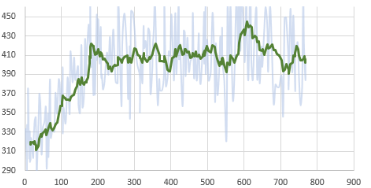}
	\caption{Kinetic energy of the air around the F1 model throughout the training duration.}
	\label{fig:ke}
\end{figure}

\begin{figure}[t!]
	\centering
	\includegraphics[width=0.7\linewidth]{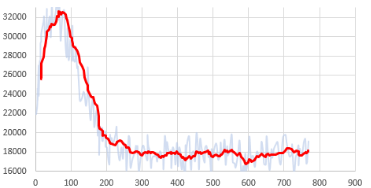}
	\caption{F1 model voxel collision count throughout the training duration.}
	\label{fig:vcc}
\end{figure}

For the training process, a batch size of 1024 samples paired with a replay buffer size of 10,240 was used. The learning rate was set to 3.0e-4, following a linear schedule. Specific to the PPO algorithm, hyperparameters include a beta value of 9.0e-3 with a constant schedule, an epsilon of 0.2 with a linear schedule, a lambda of 0.95, and 5 training epochs. The neural network architecture comprises two layers with 128 hidden units each. Memory settings encompass a sequence length of 64 and a memory size of 256. In terms of training configurations, a maximum of 5000 training steps were imposed per epoch, a time horizon of 64, and a summary update frequency of 1. Finally, the extrinsic reward signal was set to 1.0 and was subjected to a discount factor of 0.99. The training parameters are outlined in Table \ref{tab:params}, while Table \ref{tab:sim_config} lists the simulation configuration options. 

Figures \ref{fig:df}--\ref{fig:vcc} illustrate the convergence of aerodynamic metrics during the F1 model's training process. All metrics exhibit a clear improvement trend, reaching a high level of performance by approximately 200 training steps. Further training yields minimal improvements, suggesting that the model has achieved a good degree of convergence. Figure \ref{fig:collision_heatmap} illustrates the collision heatmap of the F1 car before and after the optimisation, when considering just the kinetic energy as reward signal, when considering both the kinetic energy and the drag force, and finally, when considering all metrics including the voxel collision count.

\section{Evaluation}
\label{sec:evaluation}

Table \ref{tab:drag_force} shows the average drag force experienced by different car models under various optimisation scenarios. Specifically, drag is investigated in the original prototypes, after the optimisation when considering the different optimisation objectives. Initially the F1 model had the lowest drag force as expected, but was also the one that achieved the greatest reduction, with a final improvement of approximately 14\%. Notably, the Corvette experienced the lowest improvement in drag force, achieving a reduction of approximately 11\%.

A similar trend is observed in Table \ref{tab:kinetic_energy} which presents the average kinetic energy under the different optimisation scenarios. Again, the F1 car demonstrated the highest kinetic energy, and also experienced the greatest improvement, achieving a 42\% enhancement. In this case, the other two models also experienced significant improvements, at 34\% and 23\% for the Duesen and the Corvette respectively.

\begin{table*}[t!]
    \centering
    \caption{Average Drag Force of the Vehicle.}
    \label{tab:drag_force}
    {\renewcommand{\arraystretch}{1.55}
    \begin{tabular}{llllllllll}
    \toprule
    \multicolumn{2}{c}{} & \multicolumn{2}{c}{\textbf{Optimising KE}} & \multicolumn{2}{c}{\textbf{Optimising KE + DF}} & \multicolumn{2}{c}{\textbf{Optimising KE + DF + VCC}} \\
    \midrule
    Car &  Original & Optimised & Improvement & Optimised & Improvement & Optimised & Improvement \\
    \midrule
    F1 car         & 2004.63   & 1786.41   &  -10.89\%  & 1752.57  &  -12.57\%  & 1716.85   &  -14.36\%  \\ 
    Duesen Bayern  & 2092.28   & 1798.93   &  -15.08\%  & 1871.74  &  -11.13\%  & 1791.10   &  -15.51\%  \\ 
    Corvette-C8    & 2023.85   & 1818.38   &  -10.70\%  & 1817.09  &  -10.77\%  & 1804.97   &  -11.43\%  \\
    \bottomrule
    \end{tabular}
    }
\end{table*}

\begin{table*}[t!]
    \centering
    \caption{Average Kinetic Energy of Air.}
    \label{tab:kinetic_energy}
    {\renewcommand{\arraystretch}{1.55}
    \begin{tabular}{llllllllll}
    \toprule
    \multicolumn{2}{c}{} & \multicolumn{2}{c}{\textbf{Optimising KE}} & \multicolumn{2}{c}{\textbf{Optimising KE + DF}} & \multicolumn{2}{c}{\textbf{Optimising KE + DF + VCC}} \\
    \midrule
    Car &  Original & Optimised & Improvement & Optimised & Improvement & Optimised & Improvement \\
    \midrule
    F1 car         & 283.60  & 371.41   & 30.96\%  & 391.16  & 37.93\%  & 402.78   & 42.02\%  \\ 
    Duesen Bayern  & 257.22  & 354.94   & 31.93\%  & 334.06  & 25.99\%  & 366.14   & 34.95\%  \\ 
    Corvette-C8    & 273.61  & 336.19   & 20.52\%  & 339.10  & 21.38\%  & 345.27   & 23.16\%  \\
    \bottomrule
    \end{tabular}
    }
\end{table*}
   
\begin{table*}[t!]
    \centering
    \caption{Average Voxel Collision Count.}
    \label{tab:vcc}
    {\renewcommand{\arraystretch}{1.55}
    \begin{tabular}{llllllllll}
    \toprule
    \multicolumn{2}{c}{} & \multicolumn{2}{c}{\textbf{Optimising KE}} & \multicolumn{2}{c}{\textbf{Optimising KE + DF}} & \multicolumn{2}{c}{\textbf{Optimising KE + DF + VCC}} \\
    \midrule
    Car &  Original & Optimised & Improvement & Optimised & Improvement & Optimised & Improvement \\
    \midrule
    F1 car         & 20507   & 18268   &  -11.55\%  & 17934  &  -13.39\%  & 18083   &  -12.56\%  \\ 
    Duesen Bayern  & 22122   & 18487   &  -17.90\%  & 19575  &  -12.22\%  & 17558   &  -23.00\%  \\ 
    Corvette-C8    & 20384   & 20049   &   -1.66\%  & 20018  &   -1.81\%  & 20010   &   -1.85\%  \\
    \bottomrule
    \end{tabular}
    }
\end{table*}

Finally, Table \ref{tab:vcc} illustrates the voxel collision count at each car, before and after the optimisation in the different scenarios. Interestingly, while the Duesen initially had the highest collision count, it experienced the most significant reduction after optimisation, indicating a substantial improvement of 23\%. On the contrary, the Corvette which had the lowest collision count originally, only experienced a marginal improvement, being outperformed by the other two models in the end, when it comes to this metric. A voxel collision heatmap before and after the optimisation is illustrated in Figure \ref{fig:collision_heatmap}.

While these results demonstrate the capability of DRL in achieving significant improvements in these aerodynamic metrics, it is important to acknowledge certain engineering aspects and constraints that were not explicitly incorporated into the optimisation process. Specifically, our optimisation framework does not directly account for mechanical engineering considerations such as structural integrity, manufacturability, and compliance with regulatory standards. The DRL agent operates within a virtual environment, optimising the voxel heights based on predefined objectives without explicit knowledge of physical constraints that may impact the feasibility or practicality of the resulting designs in real-world applications. Such constraints may be imposed by masking specific voxels that should not be altered during the optimisation process.

In addition, this study does not address other critical engineering factors such as thermal management, vehicle dynamics, and ergonomic considerations, which are also essential for comprehensive car design optimisation. While these factors may limit the applicability of the optimised designs in practical scenarios where holistic engineering considerations are required, they are out of the scope of this study, which focuses on the potential of DRL in the aerodynamic optimisation task. Nevertheless, the constraint imposed by the voxel height variance, ensures that significant adjustments are not made, thus maintaining the overall design and structure of the original models.

\section{Conclusion}
\label{sec:conclusion}

This paper investigated the application of DRL within voxelised car models to optimise their aerodynamic performance. Through experimentation across various optimisation scenarios, the efficacy of a PPO agent in enhancing drag force, kinetic energy, and voxel collision count across different models was examined. These metrics served as the reward signal guiding the agent to learn an optimal design policy through trial and error. The results demonstrate substantial enhancements in aerodynamic performance, achieving improvements ranging from 17\% to 38\% across the aerodynamic metrics.

The findings highlight the potential of this approach to handle the complexity of the design space and overcome challenges associated with traditional optimisation methods. In addition, they provide insights into enhancing the overall vehicle performance, fuel efficiency, and therefore environmental sustainability. This research contributes to the advancement of the understanding and application of DRL in optimising complex aerodynamic design challenges in automotive engineering, with implications for future developments in vehicle design and efficiency.

\bibliographystyle{IEEEtran}
\bibliography{ref}

\end{document}